\documentclass[journal]{IEEEtran}
\pdfoutput=1
\ifCLASSOPTIONcompsoc
  \usepackage[nocompress]{cite}
\else
  \usepackage{cite}
\fi
\ifCLASSINFOpdf
\else
\fi
\usepackage{color}
\usepackage{epstopdf}
\usepackage[pdftex]{graphicx}

\usepackage{times}
\usepackage{multirow}
\usepackage{array}
\usepackage{caption}

\usepackage{booktabs}

\usepackage{amssymb}

\expandafter\let\csname equation*\endcsname\relax
\expandafter\let\csname endequation*\endcsname\relax
\expandafter\let\csname subequations*\endcsname\relax

\usepackage{amsmath}

\usepackage{subfigure,cases}

\usepackage{float}
\usepackage{graphicx}
\graphicspath{{Figure/}}
\begin{document}
	

\title{Weather GAN: Multi-Domain Weather Translation Using Generative Adversarial Networks}

\author{
	Xuelong~Li,~\IEEEmembership{Fellow,~IEEE,}
	Kai~Kou,
	and Bin~Zhao
\IEEEcompsocitemizethanks{

\IEEEcompsocthanksitem Xuelong Li  and Bin Zhao are with School of Artificial Intelligence, Optics and Electronics (iOPEN), Northwestern Polytechnical University, Xi'an 710072, P.R. China (e-mail: xuelong\_li@nwpu.edu.cn; binzhao111@gmail.com). \emph{(Corresponding author: Bin Zhao.)}


\IEEEcompsocthanksitem Kai Kou is with School of Computer Science and School of Artificial Intelligence, Optics and Electronics (iOPEN), Northwestern Polytechnical University, Xi'an 710072, P.R. China (e-mail: koukai@mail.nwpu.edu.cn).


}}

\IEEEtitleabstractindextext{%
\begin{abstract}
	In this paper, a new task is proposed, namely, weather translation, which refers to transferring weather conditions of the image from one category to another. It is important for photographic style transfer. Although lots of approaches have been proposed in traditional image translation tasks, few of them can handle the multi-category weather translation task, since weather conditions have rich categories and highly complex semantic structures. To address this problem, we develop a multi-domain weather translation approach based on generative adversarial networks (GAN), denoted as Weather GAN, which can achieve the transferring of weather conditions among sunny, cloudy, foggy, rainy and snowy. Specifically, the weather conditions in the image are determined by various weather-cues, such as cloud, blue sky, wet ground, \emph{etc.} Therefore, it is essential for weather translation to focus the main attention on weather-cues. To this end, the generator of Weather GAN is composed of an initial translation module, an attention module and a weather-cue segmentation module. The initial translation module performs global translation during generation procedure. The weather-cue segmentation module identifies the structure and exact distribution of weather-cues. The attention module learns to focus on the interesting areas of the image while keeping other areas unaltered. The final generated result is synthesized by these three parts. This approach suppresses the distortion and deformation caused by weather translation. our approach outperforms the state-of-the-arts has been shown by a large number of experiments and evaluations.
	
	
\end{abstract}

}

\maketitle

\IEEEdisplaynontitleabstractindextext

\IEEEpeerreviewmaketitle


\section{Introduction}
\label{introduction}

\begin{figure}[tbp]
	\centering
	
	\subfigure[Sunny $ \rightarrow $ Cloudy]{
		\includegraphics[width=2.5cm, height=5cm]{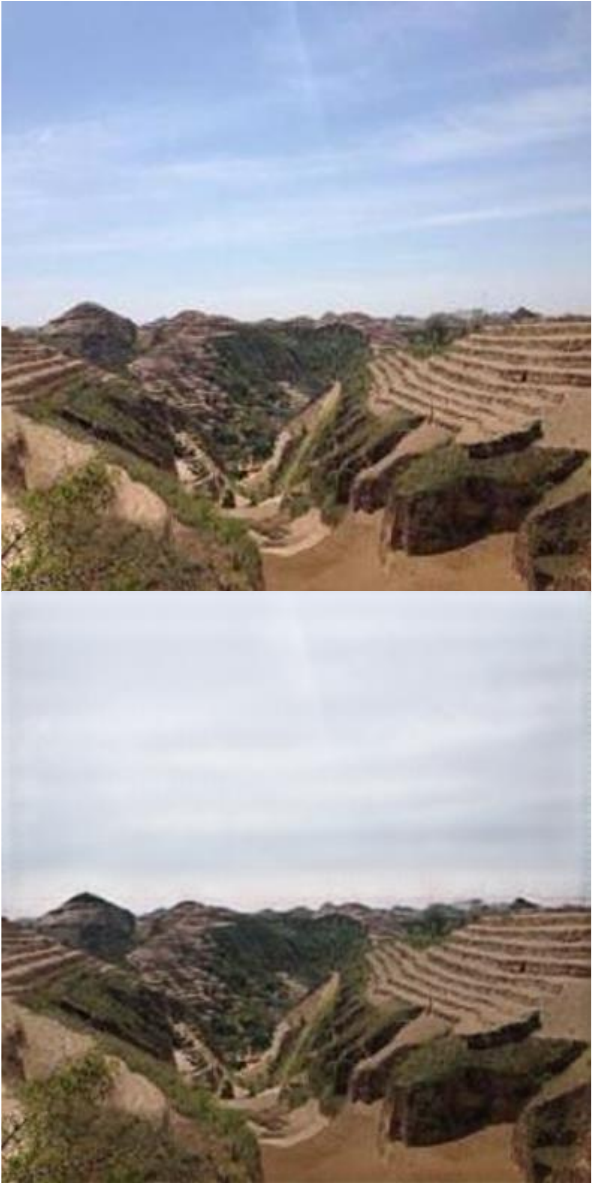}
	}%
	\subfigure[Cloudy $ \rightarrow $ Sunny]{
		\includegraphics[width=2.5cm, height=5cm]{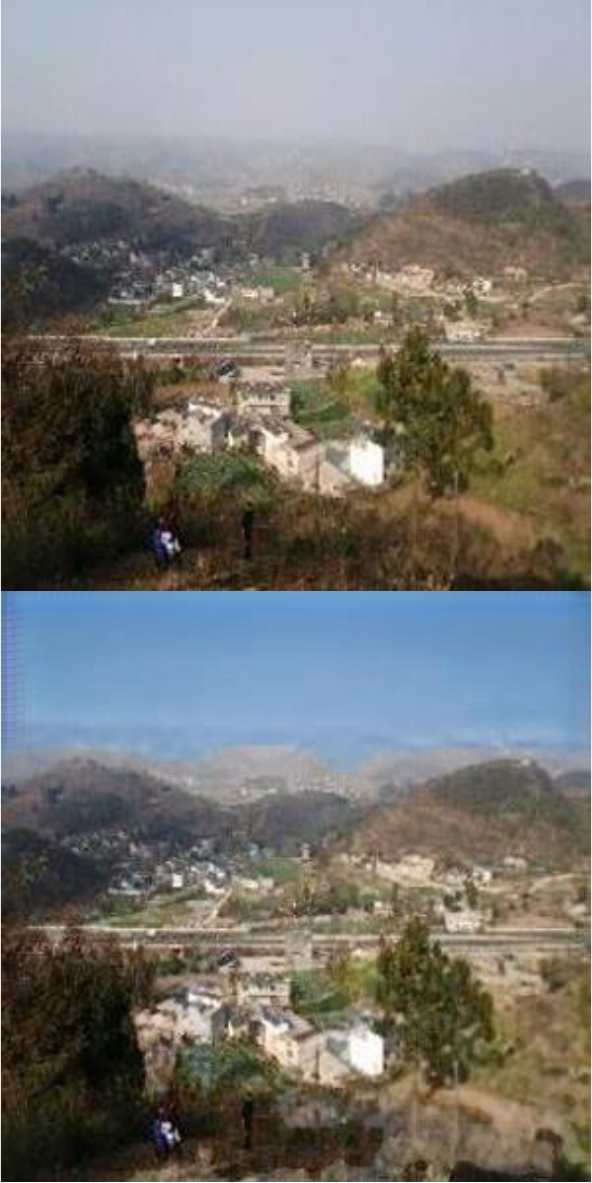}
	}%
	\subfigure[Cloudy $ \rightarrow $ Snowy]{
		\includegraphics[width=2.5cm, height=5cm]{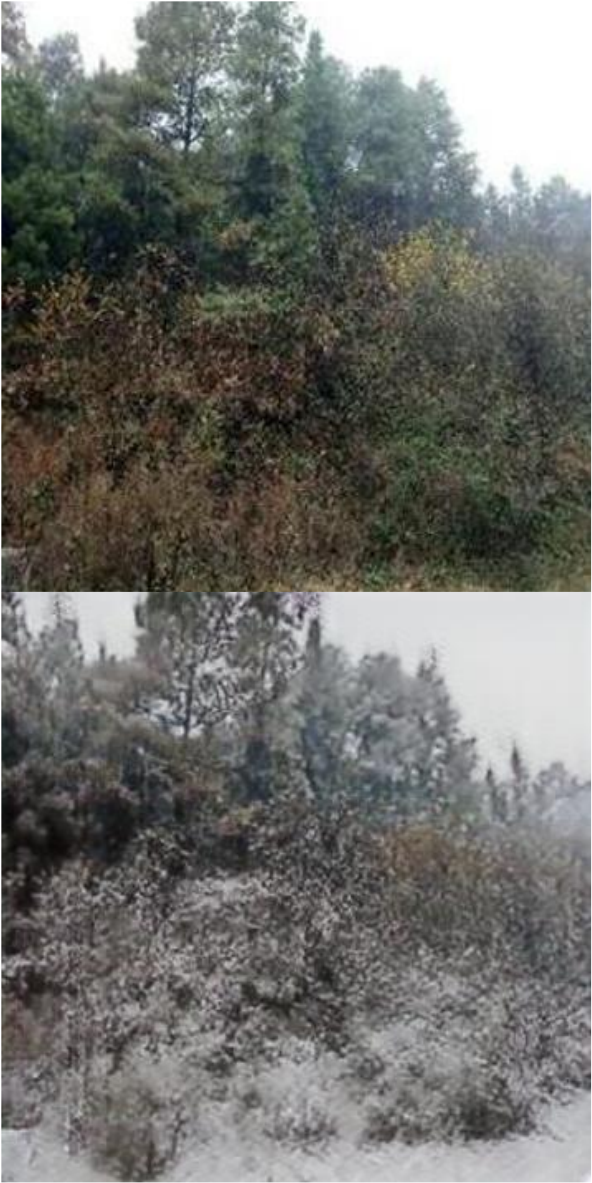}
	}%
	\caption{ Examples of weather translation.}
	\label{Fig. 1}
	\hspace{10mm}
\end{figure}

\IEEEPARstart{W}{eather} translation is a typical image translation task, which aims to transform the weather conditions in the images from one category to another. As shown in Fig. \ref{Fig. 1}, the weather conditions of the images are changed from sunny to cloudy, from cloudy to sunny, and from cloudy to snowy. It can be found that the change of weather conditions makes the image show different visual effects and artistic styles. However, the actual photographic effect is affected greatly by weather conditions \cite{DBLP:conf/cvpr/LuanPSB17}. To achieve the effect in Fig. \ref{Fig. 1}, photographers need to spend a long time finding a suitable weather condition. Therefore, we hope to develop an approach that can automatically change the weather conditions of the image. The developed approach can also benefit many other applications, such as style transfer \cite{xu2021learning, peng2020universal}, weather recognition \cite{DBLP:journals/pr/ZhaoHLLW19, DBLP:journals/ijon/ZhaoLLW18} and automatic driving \cite{DBLP:conf/kbse/ZhangZZ0K18, DBLP:conf/aaai/LiCNW17}.








Weather conditions are closely related to human production and life. There are many studies trying to handle the weather conditions in the image. Typical applications in this regard include image enhancement approaches for severe weather conditions, such as fog removal and rain removal. The early approaches have studied the formation mechanism of rain and fog, and established complex mathematical models to eliminate noise in the image. For example, the low rank model is used for rain removal \cite{DBLP:conf/iccv/ChangYZ17} and dark channel prior is used for fog removal \cite{ancuti2020day, DBLP:conf/cvpr/HeST09}. However, the real weather conditions are very complex and diverse, these traditional approaches are not capable enough to model rich information of weather conditions, therefore, it is difficult for them to achieve ideal results.

Recently, the image translation approaches based on generative adversarial networks (GAN) \cite{DBLP:conf/nips/GoodfellowPMXWOCB14} have achieved great success \cite{zhou2020cross, DBLP:conf/nips/LiuBK17, DBLP:conf/cvpr/IsolaZZE17, DBLP:conf/cvpr/AnokhinSKKKSNLS20, choi2018stargan}. These approaches aim at transforming specific features of the image at the pixel level and realizing the translation from the source domain to the target domain. Inspired by them, we try to develop an approach that can change multiple weather conditions flexibly in the image.


\subsection{Motivation and Overview}

\begin{figure}[t]
	\centering
	\includegraphics[width=0.49 \textwidth]{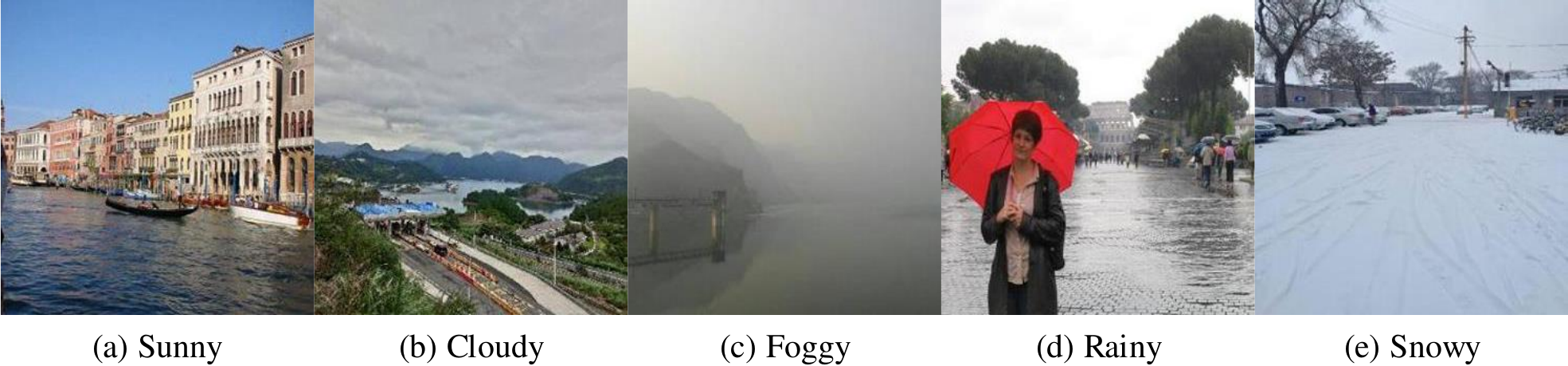}
	\caption{Examples of weather images.}
	\label{Fig. 2}
\end{figure}

Compared with general image translation tasks, weather translation faces many difficulties. On the one hand, weather conditions have multiple categories. As shown in Fig. \ref{Fig. 2}, sunny, cloudy, foggy, rainy and snowy are the five common weather conditions. Current researches mainly focus on removing noise from a single weather image \cite{jaw2020desnowgan}, such as removing fog \cite{DBLP:conf/cvpr/EnginGE18} and rain \cite{DBLP:conf/cvpr/QianTYS018}, which is a one-way image translation. However, these approaches are only applicable to specific weather conditions and cannot be adjusted for multiple weather conditions flexibly. Moreover, the results of these approaches just weaken severe weather conditions, and not change the category of the weather condition. For example, the defogging approach \cite{DBLP:conf/cvpr/EnginGE18} cannot remove rainwater. The image after removing rain only eliminates the blur and distortion caused by raindrops, but the weather condition is still rainy. Furthermore, it is useful for photography to change the weather conditions by adding noise (\emph{e.g.}, rain and snow) to image. However, few studies have paid attention to this topic. To solve these problems, we prefer to design a approach that can handle the translation of multiple weather conditions.




%

On the other hand, each weather condition has specific semantic features, such as blue sky, cloud and snow, which are also called weather-cues. Therefore, weather translation is a special image translation tasks. Ideally, rather than processing all parts of the image as a whole to change the global style, the weather translation is intended to change the weather-cues of image to another category, and keep other irrelevant areas unalter. For example, as depicted in Fig. \ref{Fig. 1}, the style of sky, tree and other areas that are highly correlated with weather condition are changed greatly, but the ground and other irrelevant areas remain basically unalter. However, weather-cues are with flexible areas in the image. For example, the blue sky in Fig. \ref{Fig. 2} (a) is only located in the upper part of the image, the fog in Fig. \ref{Fig. 2} (c) is existed in the entire of the image, and the snow in Fig. \ref{Fig. 2} (e) is only existed in the bottom of the image. It can be found that these weather-cues have no fixed structures, and even the same category of weather-cues have obvious structural differences. Therefore, the weather translation model needs to accurately predict the distribution of various weather-cues in the image \cite{DBLP:conf/nips/MejjatiRTCK18, DBLP:conf/eccv/ChenXYT18, DBLP:journals/tip/YangKWPK19, li2017locality}. Otherwise, other unrelated areas in the image may be changed incorrectly.


In this paper, we put forward the weather translation task, and develop Weather GAN to address the above problems. Specifically, the generator of Weather GAN is composed of three modules, namely, the initial translation module, the attention module, and the weather-cue segmentation module. The initial translation module is designed to take the input image as a whole to global image translation. However, the initial translation module is agnostic to the region of interest in the image, and often introduces unnecessary changes or artifacts into the result. Therefore, an attention module is designed to indicate the region of interest in the image, which enables weather translation to focus on the interesting part of the image while keeping the content of other irrelevant regions unchanged to avoid unnecessary artifacts. In addition, since the structure of weather-cues is complex, it is difficult for a single unsupervised attention module to learn the distribution of weather-cues. Therefore, a segmentation module is introduced in our approach to predict the location of weather-cues accurately. Through the integration of the weather-cue segmentation module and the attention module, the region of interest for weather translation can be predicted accurately. Finally, the generated results are synthesized by these three modules.


\subsection{Contributions}
Our contributions are summarized into the following three aspects:
 
 \begin{itemize}
 	\item The weather translation task is proposed, and a new approach, Weather GAN, is developed. This approach has important applications in photographic style transfer.
 	
 	\item An attention module is designed. It makes weather translation focus on key regions instead of randomly spreading over the whole image.
 	
 	\item A weather-cue segmentation module is introduced to indicate the exact area and structure of the weather-cues, which helps to locate the weather translation regions.
 	
 \end{itemize}

\subsection{Organization}
The following is oganized with the rest paper. Existing related works are introduced in Section II. Our proposed approach, Weather GAN, is described in detail in Section III. The experimental results and analyses are given in Section IV. Finally, the conclusion of the experiment is shown in Section V.


\begin{figure*}[h]
	\centering
	\includegraphics[width=\textwidth]{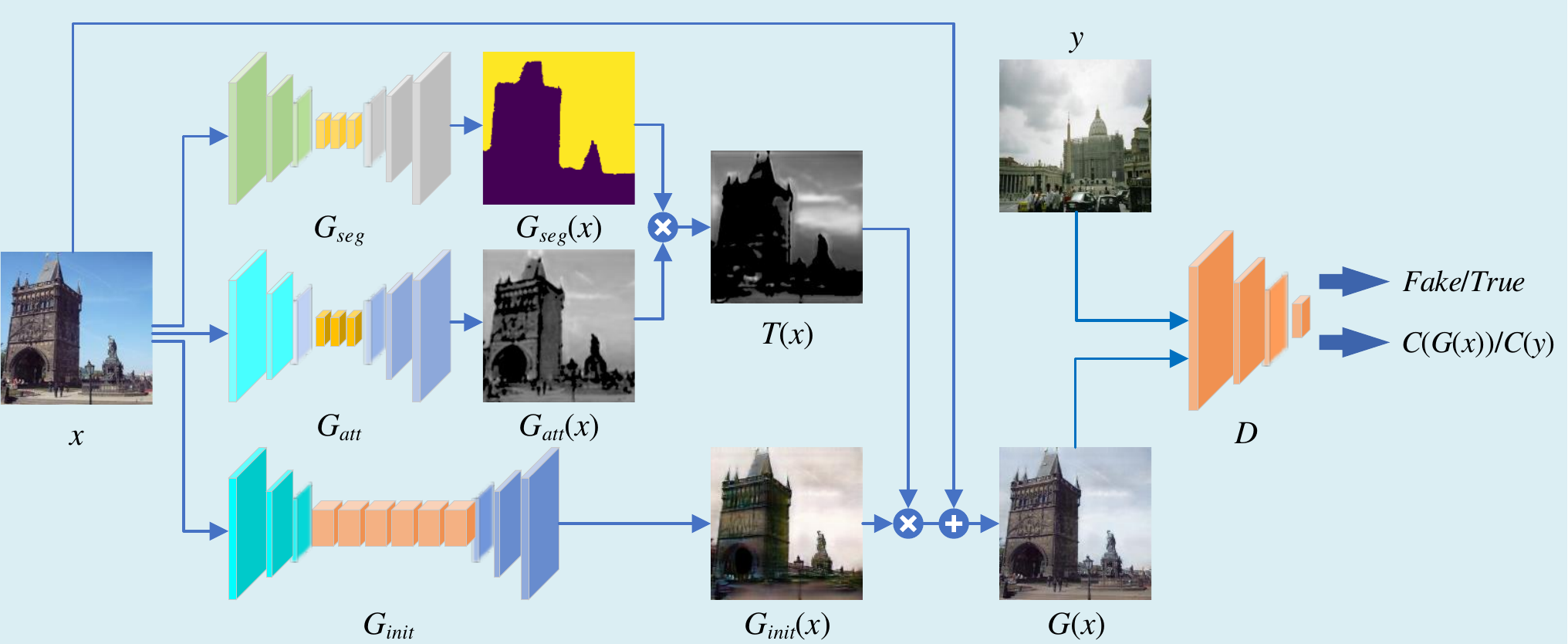}
	\caption{The model architecture of Weather GAN. The generative network $ G $ has three branches: the initial transformation module $ G_{init} $, the attention module $ G_{att} $ and weather-cue segmentation module $ G_{seg} $. The discriminator $D$ distinguishes the input image is real or fake.
	}\label{Fig. 4}
\end{figure*}
\vspace{-0.5cm}

\section{Related Works}
\label{related work}

In this section, we briefly introduce three closely related works, \emph{i.e.}, weather image research, generative adversarial networks, and image translation.

\subsection{Weather Image Research}


Weather is an integral part of our lives. Recently, understanding and processing the weather condition in the image has become one of the current research hotspots. In practical applications, it is important to recognition and handle weather conditions in the image effectively. Zhao \emph{et al}. \cite{DBLP:journals/pr/ZhaoHLLW19} propose a weather recognition approach that combines classification features and weather-cue maps, which can be adopt to recognize weather conditions in decision-making systems for autonomous driving. However, affected by various weather conditions, weather images often contain a lot of noise. In order to eliminate the influence of severe weather conditions, many researches have focused on the task of removing rain and fog in a single image. Qian \emph{et al}. \cite{DBLP:conf/cvpr/QianTYS018} introduce the attention map into the generator and discriminator to predict the position of raindrops in the image, and combines it with ground truth to remove raindrop. Engin \emph{et al}. \cite{DBLP:conf/cvpr/EnginGE18} adopt the cycle consistent structure \cite{DBLP:conf/iccv/ZhuPIE17} to eliminate fog in a single image. However, these approaches can only handle a certain category weather condition, and cannot adapt to multiple weather conditions flexibly.

Furthermore, the weather conditions give the image a certain artistic style. Some researchers attempt to transform the lighting and color in the image to realize the photography transfer style. HiDT \cite{DBLP:conf/cvpr/AnokhinSKKKSNLS20} studies the influence of illumination on imaging, and uses unlabeled scene images of different time periods to synthesize high-resolution scene images under different illuminations, realizing the conversion from day to night. Luan \emph{et al}. \cite{DBLP:conf/cvpr/LuanPSB17} propose color space constraints and realize photographic style transfer based on reference images, which can be used for the overall style transfer of the image.

\subsection{Generative Adversarial Networks}

Since the pioneering work of Goodfellow \emph{et al}. \cite{DBLP:conf/nips/GoodfellowPMXWOCB14}, the generative adversarial networks (GAN) has made impressive progress in a variety of image generation \cite{DBLP:conf/iccv/MaoLXLWS17, zou2020castle}, image editing \cite{DBLP:conf/cvpr/GatysEB16} and conditional image generation \cite{choi2018stargan, DBLP:journals/corr/abs-2011-05552}. The generator of GAN learn the distribution of training data successfully through minimax two-player game, which makes the synthetic image and real image indistinguishable.


The classic GAN is an unconditional generation model. Its input is random noise, which cause the synthesis result to be uncontrollable. To this end, the conditional GAN (cGAN) \cite{DBLP:journals/corr/MirzaO14} is developed, which can synthesize images with special categories by adding conditional information. Subsequently, Phillip \emph{et al}. propose Pix2Pix \cite{DBLP:conf/cvpr/IsolaZZE17} based on cGAN. The generator of Pix2Pix refers to the U-Net structure to improve the quality of generation result, and uses the adversarial loss and L1 loss alternately during the training process. This approach requires training data is paired and consistent, but this kind of data is usually difficult to collect. To solve this problem, CycleGAN \cite{DBLP:conf/iccv/ZhuPIE17} combines two GANs together in the network structure and proposes a cycle consistency loss training strategy, which has become a general image translation framework. At present, there are many approaches \cite{DBLP:conf/cvpr/EnginGE18, jaw2020desnowgan} trying to enhance the performance of CycleGAN by adjusting the constraint conditions and network structure.


\subsection{Image Translation}

The purpose of image translation is to map the input from the source domain to the target domain. Similar tasks include image coloring \cite{DBLP:conf/cvpr/GatysEB16}, domain adaptation \cite{DBLP:conf/cvpr/MurezKKRK18}, data augmentation \cite{DBLP:journals/corr/abs-1803-09655} and so on.

Gatys \emph{et al}. \cite{DBLP:conf/cvpr/GatysEB16} extract features from content images and style images by adopting pre-trained VGGNet \cite{DBLP:journals/corr/SimonyanZ14a} to synthesize images with specific styles, and realize the conversion of photos into Van Gogh style oil paintings. UNIT \cite{DBLP:conf/nips/LiuBK17} assumes that a shared low-dimensional latent space exists between the source and target domains, and develops an unsupervised image translation approach based on the shared hidden space. The above approaches transfer the overall style of the image, however, they may generate unrealistic results when the image contains a complex background with multiple objects. Mo \emph{et al}. \cite{DBLP:conf/iclr/MoCS19} introduce the segmentation mask of the interesting instance object in the input image, and realizes the translation between object instances, while keeping the irrelevant background unchanged. Ma \emph{et al}. \cite{DBLP:conf/cvpr/MaFCM18} decouple the local texture from the overall shape by extracting the latent space of the target data, and developed an instance-level image translation approach.

\section{Our Approach}
\label{our approach}
The architecture of the proposed approach, Weather GAN, is depicted in Fig. \ref{Fig. 4}. It consists of the weather-cue segmentation module, the attention module, the initial translation module and the discriminator, which are introduced detailedly in the following subsections.




\subsection{Model Architecture}

In our approach, considering that there are rarely few image pairs capturing the same view in different weather conditions, the weather translation task is cast as an unsupervised image translation task. Instead of transforming the image as a whole, weather translation aims to transform a specific category of weather-cues in the image into another without affecting other unrelated areas. In this process, the image translation network requires solving three equally important tasks: (1) identifying objects of the image. (2) locating the region of interest from these recognized objects. (3) the targeted region of interest is translated correctly.

Therefore, As shown in Fig. \ref{Fig. 4}, the overall architecture of Weather GAN consists of a discriminator $ D $ and a generator $ G $. $ D $ distinguishes the real image from the fake image. $ G $ transforms the input $ x $ into an image of the target domain $ Y $, which includes three branches: (1) the weather-cue segmentation module $ G_{seg} $ accurately indicates various weather-cues and generate weather-cue segmentation maps $ G_{seg}(x) $. (2) the attention module $ G_{att} $ predicts the weather translation attention map $ G_{att}(x) $, and makes the attention of weather translation focus on the region of interest instead of randomly spreading over the entire image. (3) The initial translation module $ G_ {init} $ translates the overall style of input $ x $ to generate the initial translation result $G_{init}(x)$. The generated results are synthesized by these three modules.




%


\subsection{Weather-Cue Segmentation Module}

Weather translation is to change the weather conditions of the image. The weather conditions of the image are determined by multiple weather-cues. Therefore, it is very important for weather translation to identify weather-cues in the image.

The weather-cue segmentation module $ G_{seg} $ segmentes the weather-cues in the image accurately, and generates the weather-cue segmentation maps $ G_{seg}(x) $. The size of $ G_{seg}(x) $ is the same as the input image $x$, and the elements of $ G_{seg}(x) $ represent different categories of weather-cues. Practically, the weather-cue segmentation maps mark the exact location, structure and distribution of weather-cues, which is a vital reference for identifying regions of interest in the image.


However, the distribution of weather-cues is very complex, it is a challenging task to label multiple weather-cues accurately. In this paper, the weakly supervised training strategy is adopted for the weather-cue segmentation module. The reference weather-cue segmentation maps of the training data is marked by the bounding boxes manually. The loss function is the cross entropy loss, which is defined as follows:


\begin{equation}
\begin{split}
Loss_{seg}=-\frac{1}{N}\sum_{i\in \Omega }^{}\sum_{j=1}^{N} \bigg [&a_{i,j}\log \hat{a}_{i,j}+ \\
&(1-a_{i,j})\log (1-\hat{a}_{i,j})\bigg ],
\end{split}
\end{equation} 
where $ N $ represents the size of weather-cue segmentation maps. $ a_{i,j}\in R^{N_{s}} $ is one-hot vector which the corresponding label element is one, and the other elements are zero. $ a_ {i, j} $ is the reference distribution marked by the bounding boxes, and $ \ hat {a} _ {i, j} $ represents the probability distribution predicted by the segmentation module.



\subsection{Attention Module}

Inspired by the attention of human perception \cite{rensink2000dynamic, DBLP:conf/cvpr/ZhangIESW18}, the attention module is developed to identify the region of interest for weather translation while balancing the overall style of different regions.

The attention module $G_{att}$ predicts the spatial attention map $G_{att}(x)$. The size of $G_{att}(x)$ is the same as the input $x$. The elements of $G_{att}(x)$ are continuous between [0,1], and represent the attention weights of different regions, which enables the translation attention to be focused on the region of interest and balances the overall style of the generated image.

However, the weather image contains rich background information, which may cause the attention module to predict the exact location of the region of interest incorrectly during the unsupervised training process. Therefore, $ G_{att}(x) $ and $ G_{seg}(x) $ are combined by element-wise product $ \odot $ to obtain the final translation map $ T(x) $, which is defined as follows:


\begin{equation}
	T(x)=G_{att}(x)\odot G_{seg}(x).
\end{equation}


\subsection{Initial Translation Module}


The initial translation module $ G_ {init} $ translates the global style of the input $ x $ to obtain the preliminary generated results. Firstly, the image $ x $ is input to the down-sampling layer to obtain the feature maps. The down-sampling layer has three convolution blocks, and each convolution block adopts ReLU activation and normalization \cite{DBLP:conf/icml/IoffeS15}. Then, the feature maps are extracted features through several residual block. Finally, the up-sampling layer utilizes deconvolution blocks to restore the size of the output image.


In the unpaired image translation task, $ G_ {init}(x) $ may have some irrelevant regions being converted or deformed. Therefore, in our approach, the final result of the generator is synthesized by $ G_ {init}(x) $, translation map $ T(x) $ and the original input $ x $, which is defined as follows:


\begin{equation}
G(x)=T(x)\odot G_{init}(x)+(1-T(x))\odot x.
\label{con:inventoryflow}
\end{equation}


It can be found that the translation map $ T(x) $ plays an important role in Equation \ref{con:inventoryflow}. If the elements of $ T(x) $ is replaced by all ones, then the final result $ G_(x) $ is equal to $ G_ {init}(x) $. On the contrary, If the elements of $ T(x) $ is replaced by all zeros, then the final result $ G(x) $ is equal to input $ x $. In this way, the key areas of $ G_(x) $ are closer to $ G_ {init}(x) $, while other irrelevant areas of $ G_(x) $ are closer to the input $ x $, and the edges of different areas are smooth and natural.

\subsection{Training}


In the training procedure of Weather GAN, the discriminator $D $ and the generator $G $ are optimized by the minimax game \cite{DBLP:conf/nips/GoodfellowPMXWOCB14}. In the mapping from the domain $X$ to the domain $Y$, the discriminator $D$ updates the parameters by maximizing the following objective:


\begin{equation}
\begin{split}
\mathop{max}\limits_{D} L_{adv}^{D}(D,G) =&E_{x\sim P_{data}(x)}\left [ \log (1-D(G(x)))\right ]+\\
&E_{y\sim P_{data}(y)}\left [\log D(y)\right ].
\end{split}
\end{equation}
Meanwhile, the parameters of generator $ G $ are updated by minmizing the following objective:

\begin{equation}
\mathop{min}\limits_{G} L_{adv}^{G}(D,G) =E_{y\sim P_{data}(y)}\left [ \log (1-D(G(y)))\right ].
\end{equation}
These two equations combined into:

\begin{equation}
\begin{split}
&\mathop{min}\limits_{G} \mathop{max }\limits_{D} L_{adv}(D,G,X,Y) \\
=&E_{x\sim P_{data}(x)}\left [ \log D(x)\right ]+\\
 &E_{y\sim P_{data}(y)}\left [ \log (1-D(G(y)))\right ].
\end{split}
\end{equation}

In order to achieve two-way image translation in weather GAN, similarly, in the mapping from domain $ Y $ to domain $ X $, the generator $ F $ and discriminator $ D $ are optimized by following objective:


\begin{equation}
\begin{split}
&\mathop{min}\limits_{F} \mathop{max }\limits_{D} L_{adv}(D,F,X,Y) \\
=&E_{y\sim P_{data}(y)}\left [ \log D(y)\right ]+\\
 &E_{x\sim P_{data}(x)}\left [ \log (1-D(F(x)))\right ].
\end{split}
\end{equation}

By optimizing the adversarial loss, the generator learns the distribution of the real data and can generate a realistic synthetic image. However, this constraint is not enough. Because in unsupervised training, the generator may map most data of source domain to a few samples of target domain, which makes the generated results lack of diversity. This phenomenon is also called model collapse. To solve this problem, similar to CycleGAN \cite{DBLP:conf/iccv/ZhuPIE17}, the generator is required to translate the generated results back to the source domain. Therefore, the cycle consistency loss $ L_{cycle} $ is added to our model. The cycle consistency loss of CycleGAN is L1 loss,

\begin{equation}
\begin{split}
L_{cycle}^{L1}(G,F)=&E_{x\sim P_{data}(x)}\left [ \left \| F(G(x))-x\right \|_{1}\right ]+\\
&E_{y\sim P_{data}(y)}\left [ \left \| G(F(y))-y\right \|_{1}\right ],
\end{split}
\end{equation}
which keeps the structure of the generated results, but causes the blurring of details. In order to better evaluate the semantic loss of weather translation, the perceptual loss \cite{DBLP:conf/eccv/JohnsonAF16, Ledig_2017_CVPR} is added to the model, which adopts the pre-trained VGG19 \cite{DBLP:journals/corr/SimonyanZ14a} network to extract features, and calculates the Euclidean distance of feature maps between the input and generated images,


\begin{equation}
\begin{split}
L_{cycle}^{perceptual}(G,F)=&\frac{1}{H_{j}W_{j}}\sum_{h=1}^{H_{j}}\sum_{w=1}^{W_{j}} \bigg [ \left \|\phi_{j}\left ( x\right )-\phi_{j}\left ( G(x)\right ) \right \|_{2}^{2} \\
& +\left \|\phi_{j}\left ( y\right )-\phi_{j}\left ( F(y)\right ) \right \|_{2}^{2} \bigg ],
\end{split}
\end{equation} 
where $ H_{j}, W_{j} $ are the shape of feature map, let $ \phi_{j} $ represents the $ j $th layer of VGG19. Therefore, the final cycle consistency loss is calculated by combining L1 loss and perceptual loss.


\begin{equation}
\begin{split}
L_{cycle}(G,F)=&\lambda L_{cycle}^{L1}(G,F)+\\
&(1-\lambda) L_{cycle}^{perceptual}(G,F).
\end{split}
\end{equation}

In the training process, $ L_{cycle}^{L1}(G,F) $ helps the generated results maintain the overall structure, $ L_{cycle}^{perceptual}(G,F) $ helps to generate realistic and rich detailed texture, and the hyperparameter $ \lambda $ controls the impact of these two objectives.

%


Weather conditions have clear categories \cite{DBLP:journals/pr/ZhaoHLLW19, DBLP:journals/ijon/ZhaoLLW18, DBLP:conf/mm/LiWL17}. the generated results of the existing image translation approaches have the visual effect of the target domain, but may not have the correct category of the target domain \cite{zhu2017target, zhu2017toward}. In our model, the classification loss is added to the discriminator innovatively to make the generated results with the expected target category. Let $C(x)$ represent the category of input $x$. We hope that the generated results not only have the visual effect of the target domain, but also have the correct target category, that is, $C(G(x)) = C(y)$. To this end, the classification loss $L_{class} $ is introduced and optimized by cross-entropy loss,

\begin{equation}
\begin{split}
L_{class}(G,F)=&E_{x\sim P_{data}(x)}\left ( -\log C(G(x))\right )+\\
&E_{y\sim P_{data}(y)}\left ( -\log C(F(y))\right ).
\end{split}
\end{equation}

As a consequence, the overall objective function of our model is presented as follows:

\begin{equation}
\begin{split}
L_{total}=&L_{adv}(D,G,X,Y)+L_{adv}(D,F,X,Y)+\\
&L_{cycle}(G,F)+L_{class}(G,F).
\end{split}
\end{equation}


\section{Experiments}
\label{experiments}


In this section, firstly, Weather GAN and several state-of-the-art approaches are evaluated qualitatively and quantitatively in large-scale weather image dataset. Then, the ablation study verifies the effectiveness of the weather-cue segmentation module and the attention module. Finally, user study further illustrates the effectiveness of Weather GAN.

\subsection{Experimental Details}

\subsubsection{Datasets}

\begin{figure}[t]
	\centering
	\includegraphics[width=0.49\textwidth]{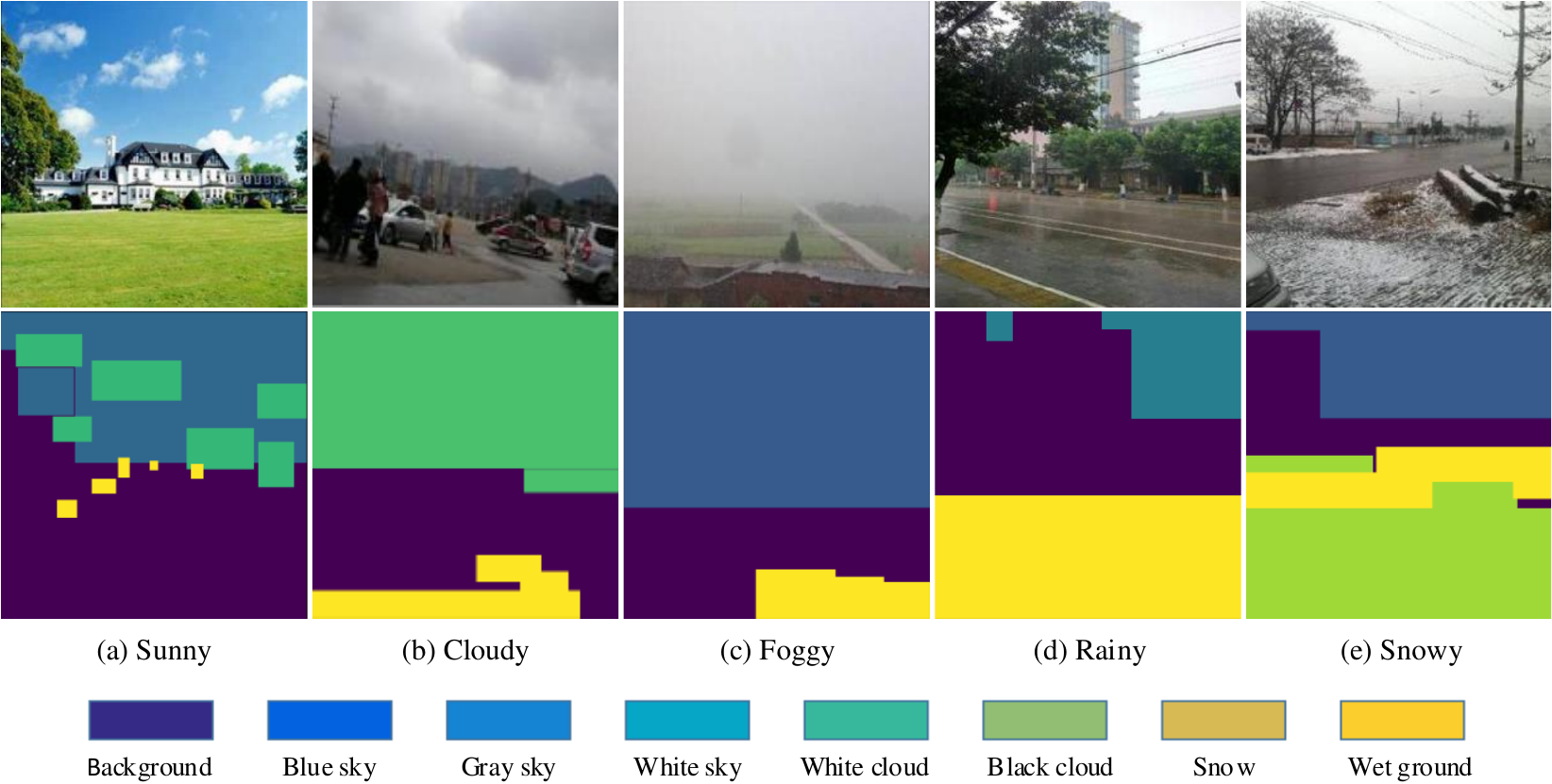}
	\caption{The five-class weather image datasets and their weather-cue segmentation maps.
	}\label{Fig. 5}
\end{figure}

The large-scale weather image dataset, Multitask Weather \cite{DBLP:journals/pr/ZhaoHLLW19}, is utilized to evaluate Weather GAN. As shown in Fig. \ref{Fig. 5}, the dataset is initially used for weather recognition task \cite{DBLP:journals/ijon/ZhaoLLW18}. Compared with other weather image datasets, it exhibits four outstanding advantages: 1) Sufficient data. It contains more than 20,000 weather images collected in natural scenes. 2) More categories. Most existing weather datasets only contain two or three weather categories \cite{DBLP:journals/pami/LuLJT17}, \cite{DBLP:journals/tip/LinLHJ17}, but this dataset contains five weather categories, including sunny, cloudy, foggy, rainy and snowy, which basically cover the common weather categories in dailylife. 3) More scenes. The data of this dataset is collected in various real outdoor natural scenes, including urban, rural, outdoor, \emph{etc.} 4) More label information. Most existing datasets only contain classification labels. However, this dataset contains weather-cue segmentation maps that marked by bounding boxes, which provides auxiliary information for weakly supervised training process.

\subsubsection{Training Strategy}

In the experiment, all images are adjusted to 300 $ \times $ 300. During the training process, the initial learning rate of the Adam solver \cite{DBLP:journals/corr/KingmaB14} is 0.0002, after 1000 iterations, the learning rate is reduced linearly in the remaining iterations. In the cycle consistency loss, the hyperparameter $ \lambda $ is 0.8 \cite{Ledig_2017_CVPR}.




\subsubsection{Baselines}

In order to verify the effectiveness of Weather GAN in the task of weather translation, three popular image translation approaches are adopted as the baselines, which are described as follows:

\begin{itemize}
	\item CycleGAN\cite{DBLP:conf/iccv/ZhuPIE17} is proposed for unpair cross-domain translation tasks. The loss functions of model are adversarial loss and cycle consistency loss.
	
	\item UNIT \cite{DBLP:conf/nips/LiuBK17} assumes that the data space of the source and target domain share the same hidden space, and combines VAE-GANs \cite{DBLP:journals/corr/KingmaW13} and cycle consistency loss \cite{DBLP:conf/iccv/ZhuPIE17} to develop an unsupervised image translation network.
	
	
	\item MUNIT \cite{DBLP:conf/eccv/HuangLBK18} assumes that the data of the source and target domain have a shared content hidden space and different style hidden spaces. Then, the results are generated by exchanging these decoupled content and style hidden spaces.
	
	
\end{itemize}

For the above baseline approaches, they are implemented with their source codes.



\subsubsection{Evaluation Metrics}

The Kernel Inception Distance (KID) \cite{DBLP:conf/iclr/BinkowskiSAG18} and Fréchet Inception Distance (FID) \cite{DBLP:conf/nips/HeuselRUNH17} are adopted to quantitatively evaluate Weather GAN and other baseline approaches. FID evaluates the similarity of two sets images by calculating the mean and variance of features, which are obtained by the Inception network \cite{DBLP:conf/cvpr/SzegedyVISW16}. KID calculates the square of the maximum average difference between the two sets of features, which represents the distribution distance between the two sets of data. In addition, KID has an unbiased estimator, which makes KID close to human perception. The lower the score of these two indicators, the closer the distribution distance between the real image and the generated image.



\begin{figure*}[!t]
	\centering
	\includegraphics[width=0.88\textwidth]{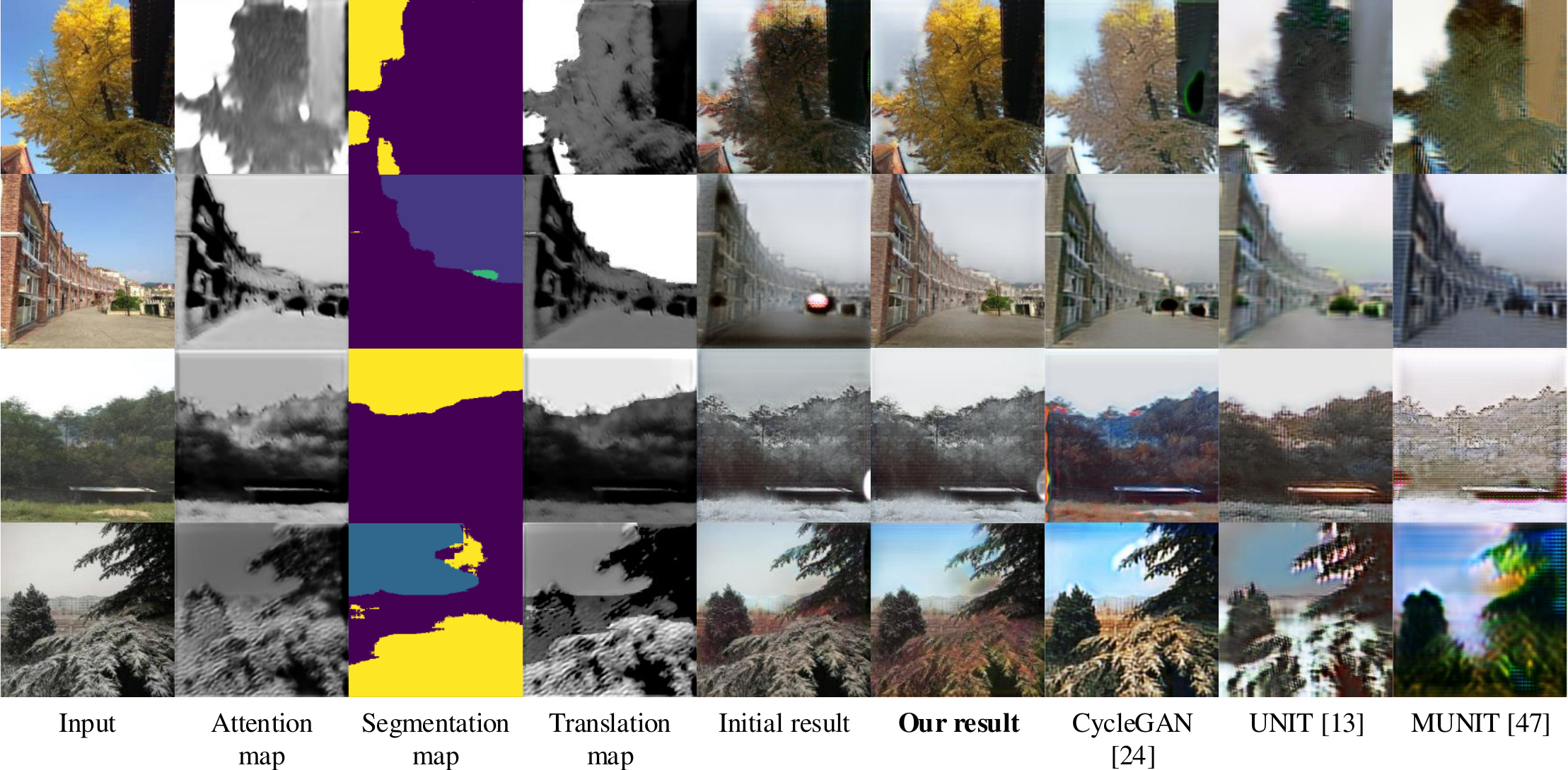}
	\caption{Weather translation results of sunny$ \rightarrow $foggy, sunny$ \rightarrow $cloudy, cloudy$ \rightarrow $snowy and snowy$ \rightarrow $sunny and comparison with CycleGAN \cite{DBLP:conf/iccv/ZhuPIE17}, UNIT \cite{DBLP:conf/nips/LiuBK17} and MUNIT \cite{DBLP:conf/eccv/HuangLBK18}.
	}\label{Fig. 6}
\end{figure*}

\begin{figure*}[!h]
	\centering
	
	\subfigure[Sunny $ \rightarrow $ Cloudy]{
		\includegraphics[width=3cm, height=6cm]{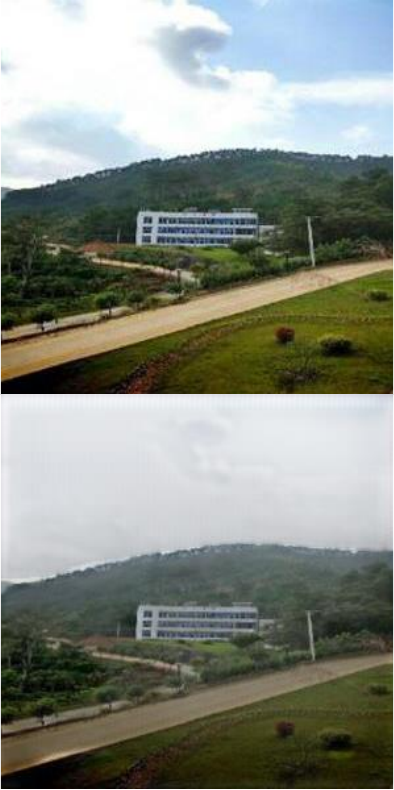}
	}%
	\subfigure[Sunny $ \rightarrow $ Foggy]{
		\includegraphics[width=3cm, height=6cm]{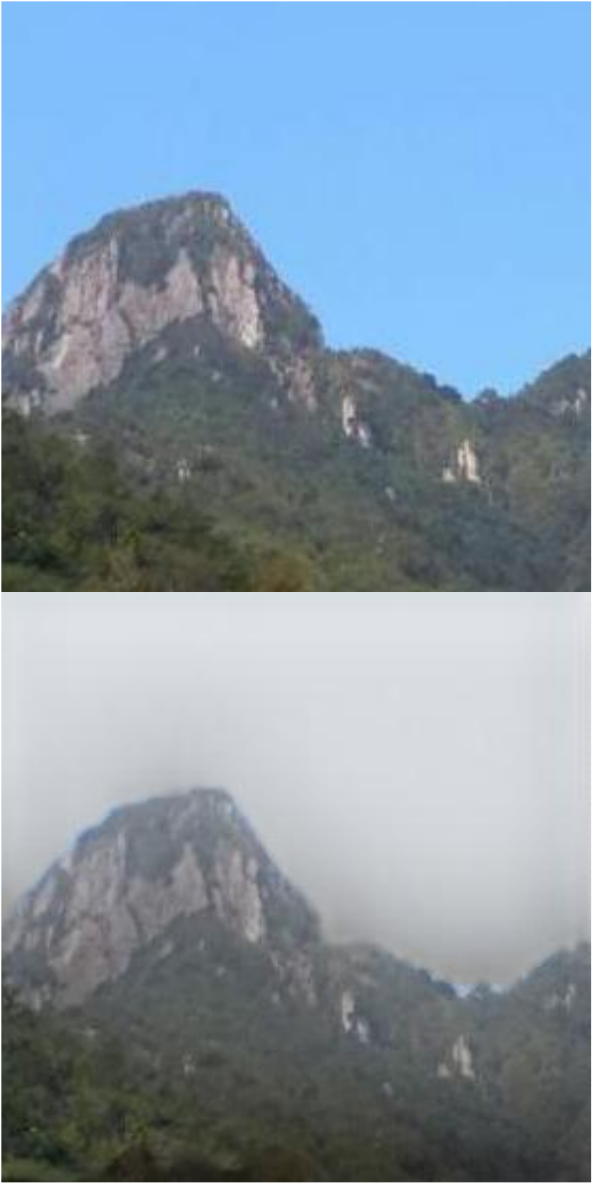}
	}%
	\subfigure[Sunny $ \rightarrow $ Snowy]{
		\includegraphics[width=3cm, height=6cm]{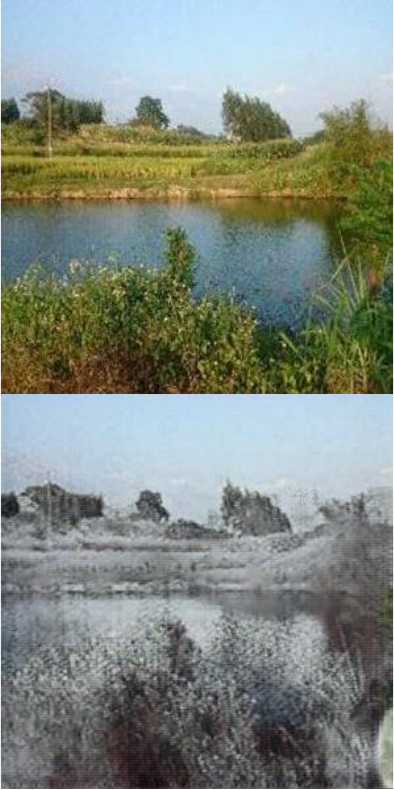}
	}%
	\subfigure[Cloudy $ \rightarrow $ Snowy]{
		\includegraphics[width=3cm, height=6cm]{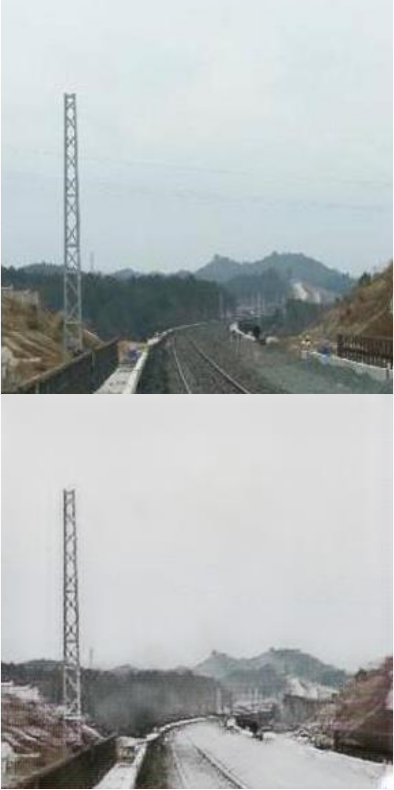}
	}%
	\subfigure[Sunny $ \rightarrow $ Rainy]{
		\includegraphics[width=3cm, height=6cm]{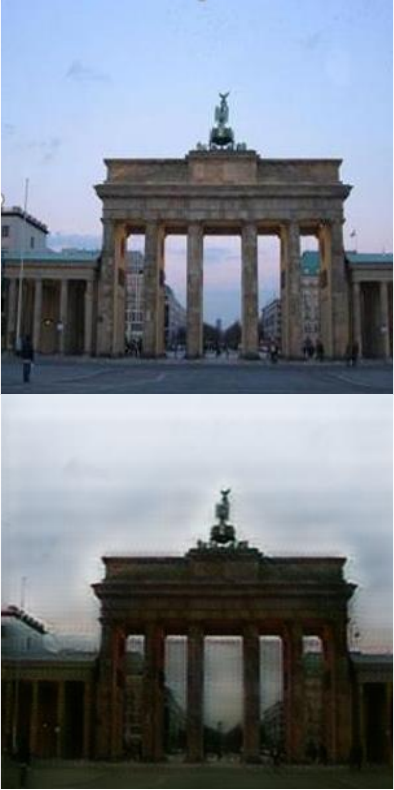}
	}%
	
	\subfigure[Cloudy $ \rightarrow $ Sunny]{
		\includegraphics[width=3cm, height=6cm]{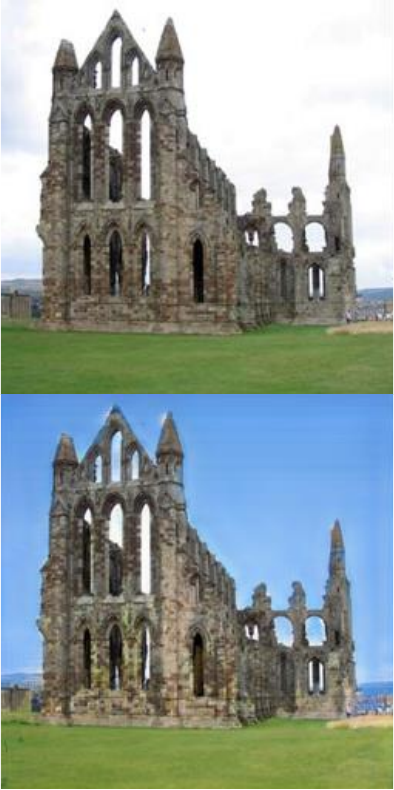}
	}%
	\subfigure[Foggy $ \rightarrow $ Sunny]{
		\includegraphics[width=3cm, height=6cm]{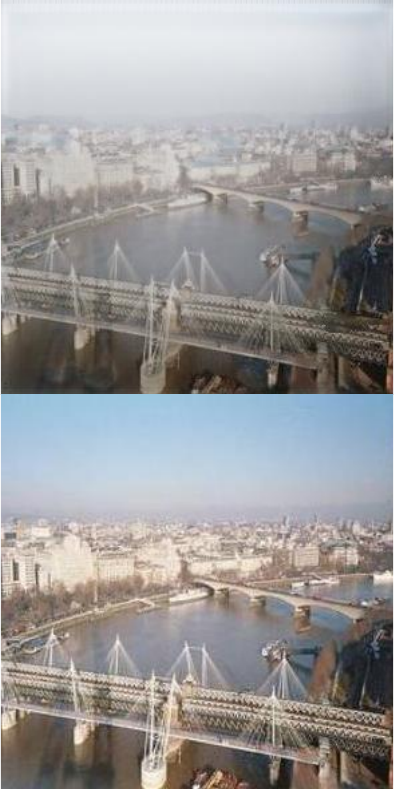}
	}%
	\subfigure[Snowy $ \rightarrow $ Sunny]{
		\includegraphics[width=3cm, height=6cm]{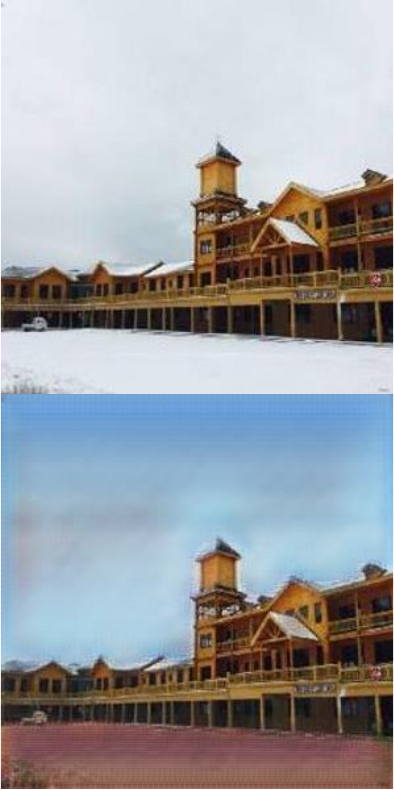}
	}%
	\subfigure[Snowy $ \rightarrow $ Cloudy]{
		\includegraphics[width=3cm, height=6cm]{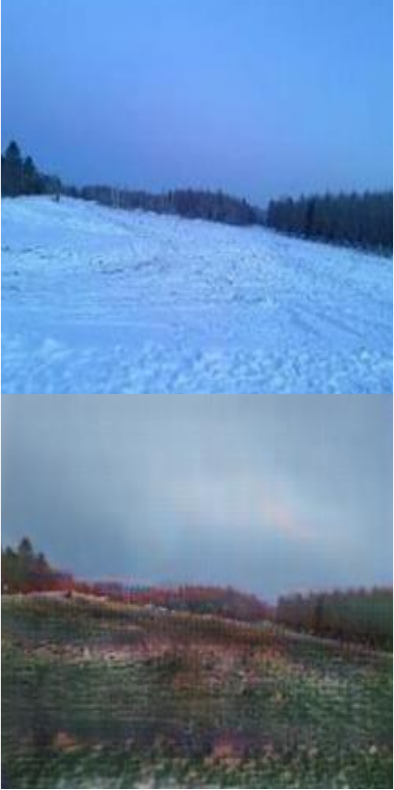}
	}%
	\subfigure[Rainy $ \rightarrow $ Sunny]{
		\includegraphics[width=3cm, height=6cm]{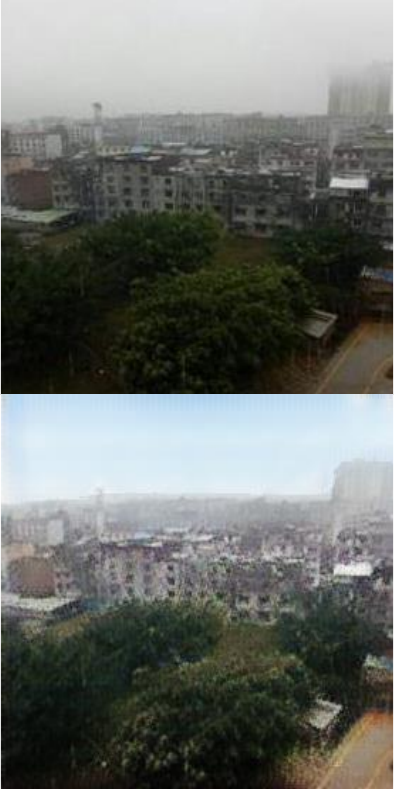}
	}%
	
	\caption{ More weather translation results generated by Weather GAN.}
	\label{Fig. 7}
\end{figure*}

\subsection{Qualitative Results}
In order to evaluate the effectiveness of the our approach qualitatively, Weather GAN and other baseline approaches are trained with various weather images. As shown in Figure 5, the input image $ x $, the initial translation result $ G_{init}(x) $, the attention map $ G_{att}(x) $, the weather-cue segmentation maps $ G_{seg}(x) $, translation map $ T(x) $ and the final result $ G(x) $ are displayed in turn, and compared with the results of CycleGAN \cite{DBLP:conf/iccv/ZhuPIE17}, UNIT \cite{DBLP:conf/nips/LiuBK17}, and MUNIT \cite{DBLP:conf/eccv/HuangLBK18}.


It can be found that our model can accurately learn the distribution of weather-cues with the help of the attention map and the weather-cue segmentation maps. Therefore, the weather translation can be performed without changing irrelevant regions. For example, in the translation from sunny to cloudy, the attention map mainly focuses on the sky, which indicates that the sky is the key distinguishing feature between sunny and cloudy. Similarly, in the translation of sunny to snowy, the attention map mainly focuses on the ground and sky.

However, the distribution of weather-cues is complex and changeable, which may cause the attention module to predict errors at the weather-cues boundary during unsupervised training. For example, the boundary in the attention map between the sky and other objects is usually unclear, which may lead to erroneous results. Therefore, the attention map and weather-cue segmentation maps need to be fused to accurately determine the location of various weather-cues of the image.

Among other competing approaches, The results of CycleGAN \cite{DBLP:conf/iccv/ZhuPIE17} appear many erroneous translation of unrelated regions, which illustrates that a single image translation network is agnostic to the region of interest. The results of UNIT \cite{DBLP:conf/nips/LiuBK17}, and MUNIT \cite{DBLP:conf/eccv/HuangLBK18} have a lot of blur and distortion, which is mainly due to the huge semantic gap between various weather conditions, and the image contains a lot of other objects. Therefore, the shared latent space assumption is difficult to adapt to multi-domain weather image translation task.

\begin{figure}[h]
	\centering
	\includegraphics[width=0.5\textwidth]{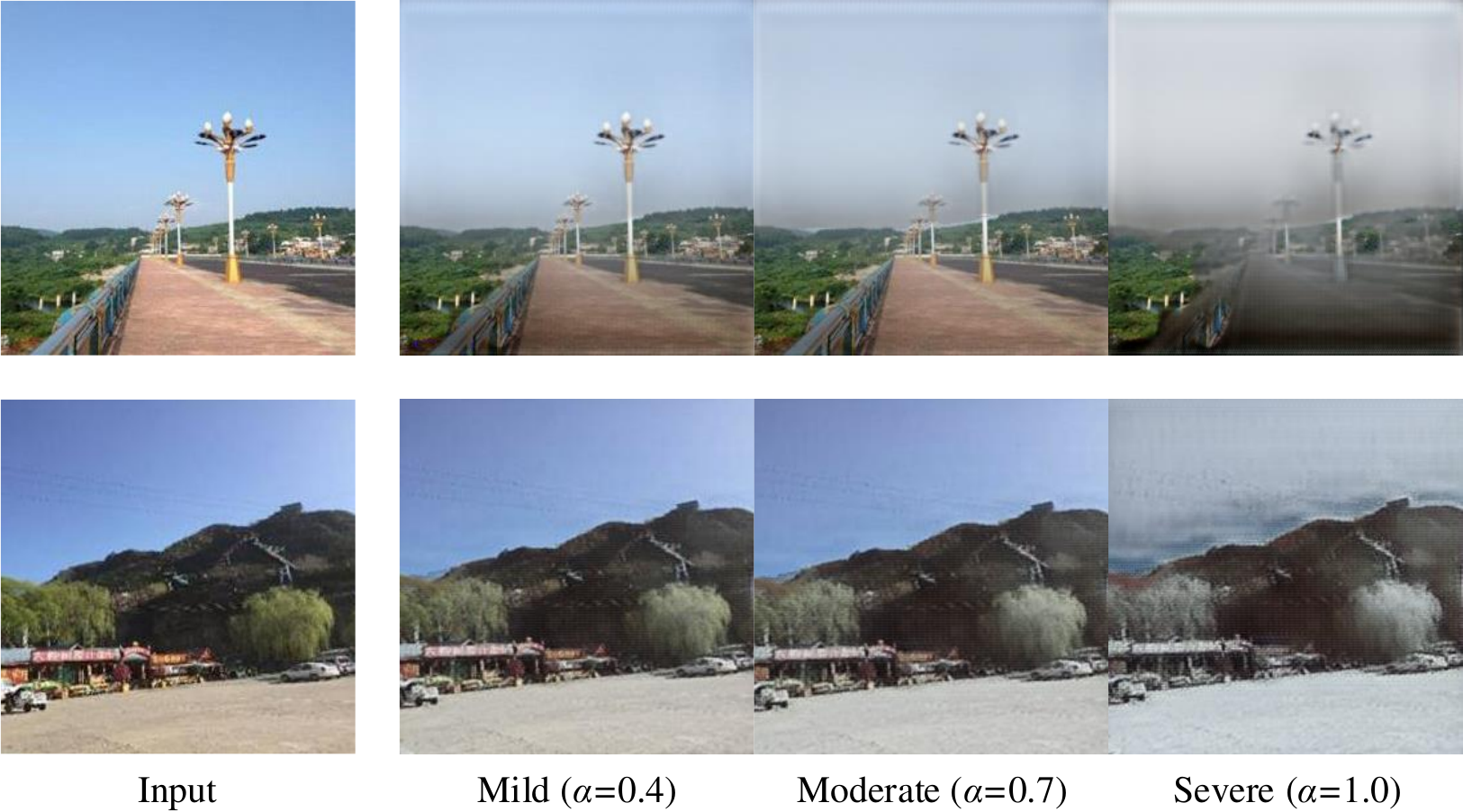}
	\caption{Weather translation results generated by Weather GAN with different intensity weather conditions. From top to bottom: sunny$ \rightarrow $foggy, sunny$ \rightarrow $snowy, and from left to right, the intensity of weather conditions gradually increases.
	}\label{Fig. 8}
\end{figure}

Fig. \ref{Fig. 7} shows more weather translation results of Weather GAN, whcih indicates that Weather GAN can realize the translation of various weather conditions, and the visual effect is smooth and natural. Moreover, as shown in Fig. \ref{Fig. 8}, in the process of synthesizing the final results, Weather GAN can generate results with various intensities of weather conditions, by adjusting the weights $ \alpha $ of input $ x $ and the initial translation result $ G_{init}(x) $,

\begin{equation}
G(x)=\alpha T(x)\odot G_{init}(x)+(1-\alpha T(x))\odot x,
\end{equation}
which greatly improves the diversity of the final synthesized image.




\begin{table*}[h]
	
	\centering
	\caption{The KID \cite{DBLP:conf/iclr/BinkowskiSAG18} and FID \cite{DBLP:conf/nips/HeuselRUNH17} scores of translation results for multi-domain weather conditions, lower is better.}\label{Table1}
	\renewcommand\arraystretch{1.5}
	\begin{tabular}{m{3cm}<{\centering}||m{1.3cm}<{\centering}|m{1.3cm}<{\centering}|m{1.3cm}<{\centering}|m{1.3cm}<{\centering}|m{1.3cm}<{\centering}|m{1.3cm}<{\centering}|m{1.3cm}<{\centering}|m{1.3cm}<{\centering}}
		\toprule[1.5pt]
		\multirow{2}*{Weather Class}      &\multicolumn{2}{c|}{\textbf{Weather GAN}}    &\multicolumn{2}{c|}{CycleGAN \cite{DBLP:conf/iccv/ZhuPIE17}}  &\multicolumn{2}{c|}{UNIT \cite{DBLP:conf/nips/LiuBK17}}    &\multicolumn{2}{c}{MUNIT \cite{DBLP:conf/eccv/HuangLBK18}}\\ \cline{2-9}
		&FID  &KID                     &FID  &KID                   &FID  &KID                     &FID  &KID  \\ 
		\midrule[1.0pt]
		sunny$ \leftrightarrow $cloudy          &86 &0.010                    &95 &0.027                 &99 &0.022                    &101 &0.023 \\
		sunny$ \leftrightarrow $foggy           &98 &0.016                    &105 &0.027                &103 &0.029                   &108 &0.035 \\
		sunny$ \leftrightarrow $snowy            &100 &0.025                   &108 &0.015                &105 &0.023                   &115 &0.042 \\
		sunny$ \leftrightarrow $rainy           &95 &0.020                    &110 &0.038                &112 &0.021                   &100 &0.022 \\
		cloudy$\leftrightarrow $snowy            &110 &0.030                   &137 &0.033                &124 &0.039                   &120 &0.047 \\
		cloudy$\leftrightarrow $rainy            &97 &0.022                    &117 &0.039                &108 &0.047                   &102 &0.023 \\
		cloudy$\leftrightarrow $foggy           &95 &0.018                    &99 &0.020                 &103 &0.029                   &105 &0.032 \\
		foggy$ \leftrightarrow $rainy            &107 &0.028                   &130 &0.045                &109 &0.033                   &130 &0.060 \\
		snowy$ \leftrightarrow $foggy           &105 &0.017                   &126 &0.024                &114 &0.027                   &127 &0.035 \\
		snowy$ \leftrightarrow $rainy            &121 &0.035                   &140 &0.050                &139 &0.045                   &145 &0.079 \\
		\bottomrule[1.5pt]
	\end{tabular}
\end{table*}

\subsection{Quantitative Results}

To validate the performance of the Weather GAN, the generated results are calculated the KID and FID scores. We randomly selected 500 generated result images from each domain as testing data to ensure the consistency of the calculation results.

The results of the evaluation indicators are shown in Table \ref{Table1}. It can be found that Weather GAN obtains the lowest KID and FID scores among all competing approaches. CycleGAN \cite{DBLP:conf/iccv/ZhuPIE17} is the second best approach, and UNIT \cite{DBLP:conf/nips/LiuBK17} and MUNIT \cite{DBLP:conf/eccv/HuangLBK18} have their own advantages in different categories of results. In addition, in the same model, the scores of image translation results in various categories are different, which indicates that the difficulty of image translation in various categories weather conditions is different, which is also consistent with our qualitative results.

\begin{table}[h]
	\centering
	\caption{The classification results on the Weather GAN generated image.}\label{Table2}
	\renewcommand\arraystretch{1.5}
	\begin{tabular}{m{1.7cm}<{\centering}||m{1.6cm}<{\centering}|m{1.2cm}<{\centering}|m{1cm}<{\centering}|m{1.2cm}<{\centering}}
		\toprule[1.5pt]
		Weather Class  &\textbf{Weather GAN} &CycleGAN \cite{DBLP:conf/iccv/ZhuPIE17} &UNIT \cite{DBLP:conf/nips/LiuBK17} &MUNIT \cite{DBLP:conf/eccv/HuangLBK18}  \\
		\midrule[1.0pt]
		sunny$ \leftrightarrow $cloudy  &87$\%$ &64$\%$ &65$\%$ &68$\%$\\
		sunny$ \leftrightarrow $foggy   &82$\%$ &70$\%$ &59$\%$ &48$\%$\\
		sunny$ \leftrightarrow $snowy    &75$\%$ &67$\%$ &50$\%$ &55$\%$\\
		sunny$ \leftrightarrow $rainy    &63$\%$ &56$\%$ &36$\%$ &42$\%$\\
		cloudy$\leftrightarrow $snowy    &76$\%$ &63$\%$ &60$\%$ &58$\%$\\
		cloudy$\leftrightarrow $rainy    &62$\%$ &48$\%$ &52$\%$ &58$\%$\\
		cloudy$\leftrightarrow $foggy   &70$\%$ &65$\%$ &68$\%$ &61$\%$\\
		foggy$ \leftrightarrow $rainy    &50$\%$ &44$\%$ &35$\%$ &37$\%$\\
		snowy$ \leftrightarrow $foggy   &57$\%$ &48$\%$ &45$\%$ &42$\%$\\
		snowy$ \leftrightarrow $rainy    &44$\%$ &34$\%$ &35$\%$ &32$\%$\\
		\bottomrule[1.5pt]
	\end{tabular}
\end{table}


In addition, The purpose of Weather GAN is to change the category of the weather conditions in the image from one to another. In order to evaluate whether the generated results have the correct target weather condition category, the generated results are tested by the latest weather recognition approach \cite{DBLP:journals/pr/ZhaoHLLW19}, which can recognize five weather condition categories. Firstly, the weather classification approach is pre-trained on our training dataset \cite{DBLP:journals/pr/ZhaoHLLW19}, and then, the generated results are adopted for classification experiments. 

As shown in Table \ref{Table2}, most of the generated results of Weather GAN can be identified as the correct target category, and the highest accuracy rate is 87$\%$ and the lowest accuracy rate is 45$\%$. Compared with other baseline approaches, Weather GAN has obvious advantages. Similar to KID and FID, the classification accuracy of various categories also has a large gap, which indicates that information loss is caused in the weather translation process.

\begin{figure}[!h]
	\centering
	\includegraphics[width=0.5\textwidth]{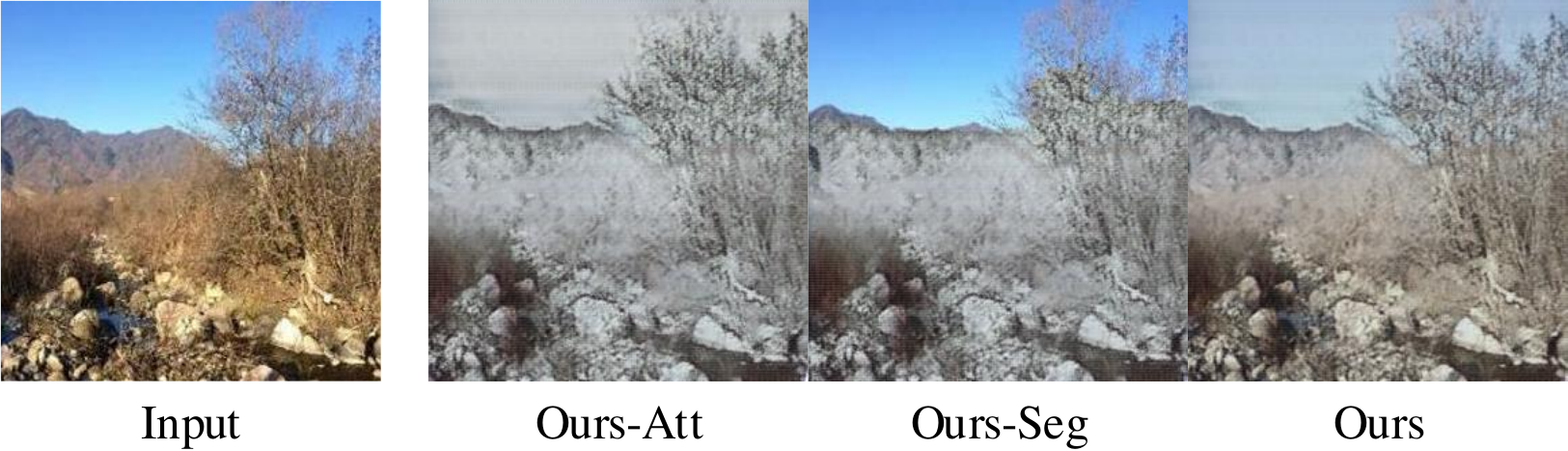}
	\caption{The influence of different modules on the synthesis image results.}
	\label{Fig. 9}
\end{figure}

\subsection{Ablation Study}

In order to better understand the impact of each part of the Weather GAN, we removed the attention module and the weather-cue segmentation module, respectively. The results in Fig. \ref{Fig. 9} turn out that a single attention module has some error in predicting the target region in the translation from sunny to snowy, which cause irrelevant regions to be translated. Meanwhile, the result of only using the weather-cue segmentation module is equivalent to image splicing, and the overall visual effect is unnatural, especially, there is no smoothness at the boundary between the sky and the mountain. Therefore, Weather GAN guides the final image translation by fusing these two modules, which further shows that our multiple module fusion mechanism can improve the visual effect of the generated results.

\begin{figure}[t]
	\centering
	\includegraphics[width=0.5\textwidth]{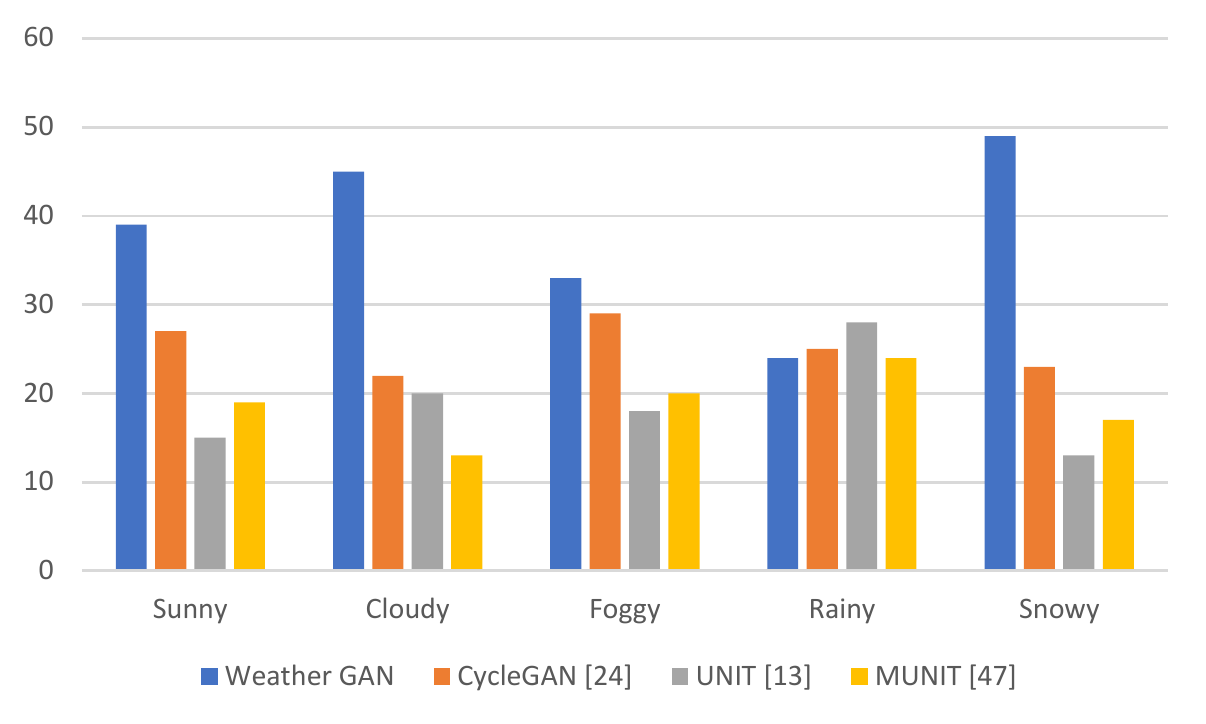}
	\caption{The user study on weather translation results.}
	\label{Fig. 10}
\end{figure}

\subsection{User Study}

In order to further evaluate the performance of Weather GAN more rigorously. user study is conducted to evaluate the results generated by various approaches. 10 subjects are employed to provide feedbacks, and each subject get 10 sets of images for comparison. Each set contains four generated result images of Weather GAN and other baseline approaches, and these data are arranged in random order. Then, the subjects are asked to sort the images in each set from high to low visual quality. Fig. \ref{Fig. 10} shows the results of the user study. It can be seen that the most of our results are better than other compared approaches, which is consistent with the results of quantitative and qualitative experiments.

\subsection{Failure Cases}

\begin{figure}[h]
	\centering
	\includegraphics[width=0.49\textwidth]{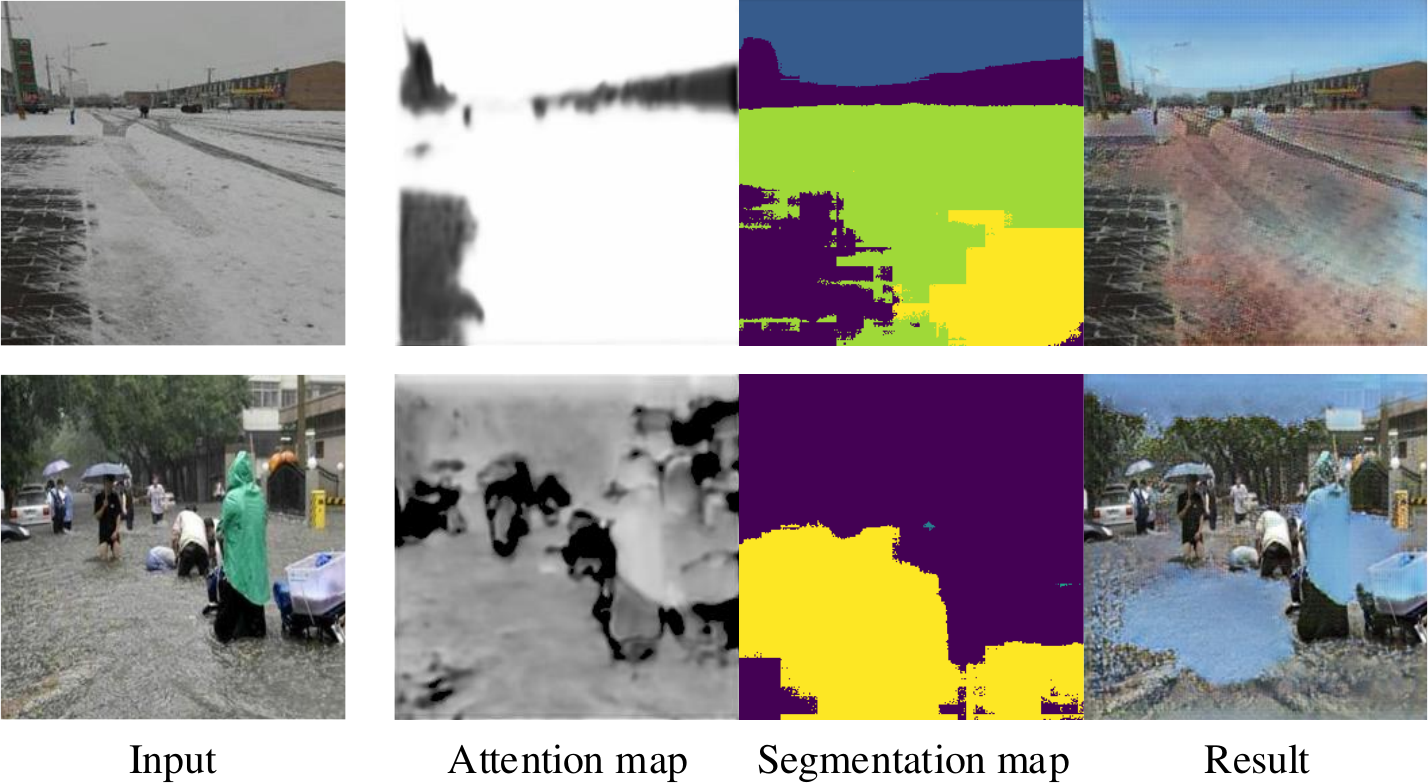}
	\caption{Failure cases of Weather GAN. From top to bottom: snowy$ \rightarrow $sunny, rainy$ \rightarrow $sunny.}
	\label{Fig. 11}
\end{figure}

Although Weather GAN can generate realistic results in the multiple weather conditions, it still has limitations in some cases. As shown in Fig. \ref{Fig. 11}, affected by bad weather conditions, such as snow and fog, many areas of the image are covered and deformed. Due to the lack of other additional information, these defects cannot be restored to the state before being covered during the weather translation procedure. Therefore, even if our attention map and weather-cue segmentation maps accurately indicate the key areas of weather-cues, the generated results still have some faultiness, such as blur and color distortion.

\section{Conclusion}
\label{Conclusion}
In this paper, a novel weather translation framework, Weather GAN, is developed to change the weather conditions in natural scene image. In order to identify the weather-cues of the image, a weather-cue segmentation module is introduced. In addition, inspired by the attention mechanism, an attention module is added to focus the attention of image translation on the region of interest and eliminate unnecessary changes or artifacts. Finally, Weather GAN can focus the main attention on the weather-cues and balance the overall style, by fusing the attention module and the weather-cue segmentation module. Qualitative and quantitive evaluations show that Weather GAN can provide visually realistic weather translation results. Furthermore, although Weather GAN is proposed for weather translation task, it can be transferred to other multi-domain image translation tasks with complex background transformations. In future work, we hope to explore image translation approach that can handle multiple natural scene transitions.

\bibliographystyle{IEEEtran}
\bibliography{main}
  
\vspace{0.5cm}
\begin{IEEEbiographynophoto}
	{Xuelong Li} is a full professor with School of Artificial Intelligence, Optics and Electronics (iOPEN), Northwestern Polytechnical University, Xi’an 710072, P. R. China.
\end{IEEEbiographynophoto}
\vspace{-1cm}
\begin{IEEEbiography}
	[{\includegraphics[width=1in,height=1.25in,clip,keepaspectratio]{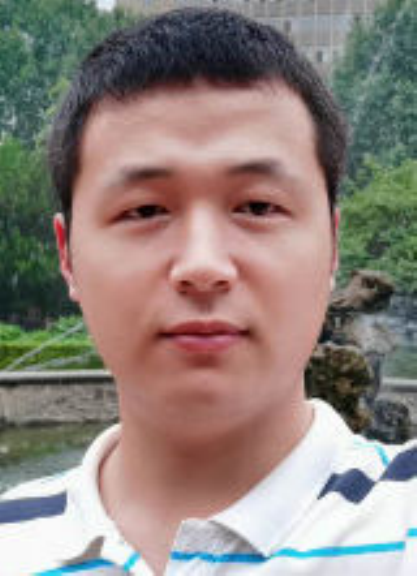}}]
	{Kai Kou} is currently pursuing the Ph.D. degree with the School of Computer Science and School of Artificial Intelligence, Optics and Electronics (iOPEN), Northwestern Polytechnical University, Xi’an, China. His research interests include generative image modeling, computer vision and machine learning.
\end{IEEEbiography}
\vspace{-1cm}
\begin{IEEEbiography}
	[{\includegraphics[width=1in,height=1.25in,clip,keepaspectratio]{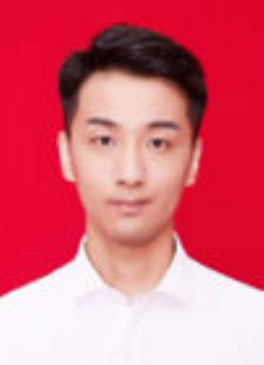}}]
	{Bin Zhao} is a doctor with the School of Artificial Intelligence, Optics and Electronics (iOPEN), Northwestern Polytechnical University, Xi’an 710072, P. R. China. His research interests are introducing physics models and cognitive science to artificial intelligence.
\end{IEEEbiography}

\end{document}